\documentclass{article}

\usepackage[preprint,nonatbib]{neurips_2022}

\usepackage{microtype}
\usepackage{graphicx}
\usepackage{subfigure}
\usepackage{multibib}

\usepackage{amsmath}
\usepackage{amssymb}
\usepackage{mathtools}
\usepackage{amsthm}
\usepackage{xspace}
\usepackage{pifont}
\newcommand{\cmark}{\ding{51}}%
\newcommand{\xmark}{\ding{55}}%

\theoremstyle{plain}

\theoremstyle{definition}

\theoremstyle{remark}

\usepackage[textsize=tiny]{todonotes}

\usepackage[utf8]{inputenc} 
\usepackage[T1]{fontenc}    
\usepackage{hyperref}       
\usepackage{url}            
\usepackage{booktabs}       
\usepackage{amsfonts}       
\usepackage{nicefrac}       
\usepackage{microtype}      
\usepackage{xcolor}         

\usepackage[capitalize,noabbrev]{cleveref}

\title{HiP: Hierarchical Perceiver}

\author{
João Carreira$^{1}$ \quad Skanda Koppula$^{1}$ \quad Daniel Zoran$^{1}$ \quad Adria Recasens$^{1}$ \quad Catalin Ionescu$^{1}$ \\
\textbf{Olivier H\'enaff}$^{1}$ \quad \textbf{Evan Shelhamer}$^{1}$ \quad \textbf{Relja Arandjelovi\'c}$^{1}$\ \quad \textbf{Matt Botvinick}$^{1}$ \\ 
\quad \textbf{Oriol Vinyals}$^{1}$ \textbf{Karen Simonyan}$^{2}$\thanks{Work done while at DeepMind} \quad \textbf{Andrew Zisserman}$^{1,3}$ \quad \textbf{Andrew Jaegle}$^{1}$ \\
$^{1}$DeepMind \quad $^{2}$Inflection AI\quad $^{3}$University of Oxford\\}

\newcommand{\ourmodel}{HiP\xspace}

\newcommand{\X}{\mathbf{X}}
\newcommand{\Z}{\mathbf{Z}}
\newcommand{\Y}{\mathbf{Y}}

\begin{document}

\maketitle

\vspace{-5mm}
\begin{abstract}

General perception systems such as Perceivers can process arbitrary modalities in any combination and are able to handle up to a few hundred thousand inputs. They achieve this generality by using exclusively global attention operations. This however hinders them from scaling up to the inputs sizes required to process raw high-resolution images or video. In this paper, we show that some degree of locality can be introduced back into these models, greatly improving their efficiency while preserving their generality. To scale them further, we introduce a self-supervised approach that enables learning dense low-dimensional positional embeddings for very large signals. We call the resulting model a Hierarchical Perceiver (\ourmodel). 
In sum our contributions are: 1) scaling Perceiver-type models to raw high-resolution images and audio+video, 2) showing the feasibility of learning 1M+ positional embeddings from scratch using masked auto-encoding, 3) demonstrating competitive performance on raw data from ImageNet, AudioSet, PASCAL VOC, ModelNet40 and Kinetics datasets with the same exact, unchanged model and without specialized preprocessing or any tokenization.

\end{abstract}

\section{Introduction}
\label{introduction}

\textit{Perceiver}~\cite{jaegle2021perceiver,jaegle2021perceiverio} is a recently proposed model that produces competitive results in classic vision tasks like optical flow and ImageNet classification, in text benchmarks like GLUE, in multimodal domains such as audio-visual recognition and auto-encoding, on point cloud data or even in StarCraft agents. All with the same basic architecture and without the need for any data preprocessing for up to~100k inputs / outputs.

Perceivers build upon Transformers~\cite{vaswani2017attention} but introduce an efficient global attention mechanism that allows them to handle much larger input/output spaces.
They operate on a matrix of flattened inputs -- where rows correspond to pixels, raw audio time steps, characters, etc., and columns correspond to features, such as RGB values. Perceivers produce a flat matrix of outputs controlled by an input query matrix~\cite{jaegle2021perceiverio}. Prior information, such as spatial coordinates, is featurized -- typically using Fourier features for images and audio data -- and concatenated with the inputs and the queries.

\begin{figure}[t]
\centering
\includegraphics[width=9cm]{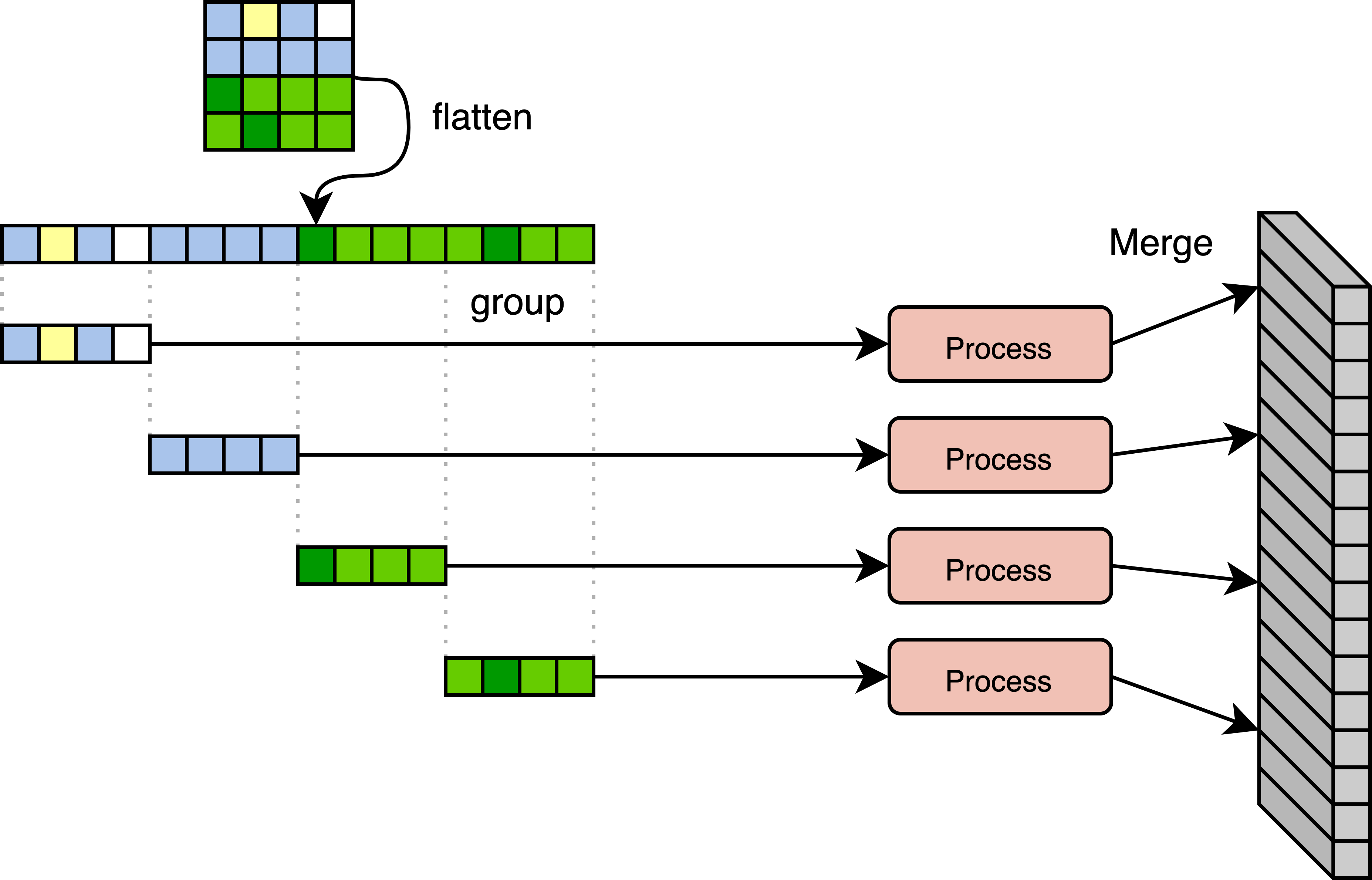}
\caption{Flattening preserves some locality information in data such as images and video: splitting a flattened input array uniformly into groups preserves useful local structure within each group.
We create local receptive fields in Perceiver models, significantly increasing their efficiency without sacrificing their generality, by introducing a sequence of hierarchical splitting-processing-merging blocks.}
\label{fig:flattening}
\vspace{-15pt}
\end{figure}

In this paper we introduce Hierarchical Perceiver (\ourmodel), a model that builds upon Perceivers, but is faster, scales to even more inputs/outputs, reduces the need for input engineering and is also more accurate. These improvements stem from two ideas. First, we build back locality into the architecture while preserving its modality-independence, by leveraging the observation that intrinsic locality in input and target output signals is often preserved after even simple, generic flattening operations, as illustrated in Figure~\ref{fig:flattening}. Second, we show that random masked auto-encoding (MAE) enables us to learn low-dimensional positional embeddings for high-resolution signals such as images or audio, doing away with the need for externally-defined, hand-designed Fourier embeddings, which are also computationally expensive. Previous papers also experimented with learned positional embeddings, but reported poor results when just training for supervised classification tasks.

We describe HiP's architecture in Section~\ref{architecture}, then analyze and ablate the model on the ImageNet dataset, showing that it is faster and more accurate than its predecessors. We also show that learned low-dimensional positional embeddings are impoverished and underperform if trained with a supervised classification loss, but are rich and slightly better than high-dimensional Fourier embeddings if trained with random pixel masked auto-encoding. We also report competitive results on AudioSet using raw audio and video, without patching or convolutions (we do not use those in this paper). Finally, we report results for semantic segmentation on PASCAL VOC and analyze the versatility of the model on ModelNet40 point cloud classification. The paper concludes with a discussion of connections of this work to other areas and current limitations.

\section{Related Work}
\label{related_work}

Computer perception is a research area that seeks to develop flexible models that can solve tasks in images, video, audio, and many other modalities. We outline work on computer perception that feeds into the Hierarchical Perceiver.

\noindent \textbf{Attention and general-purpose architectures.} Transformers~\cite{vaswani2017attention} provide a convenient building block for general-purpose architectures because they do not make strong assumptions about the structure of the data, unlike ConvNets~\cite{lecun1998gradient}. Transformers, however, have a well known quadratic growth in complexity with the number of inputs. ViT-based models~\cite{dosovitskiy2020image} have been applied to multimodal inputs~\cite{akbari2021vatt,nagrani2021attention, likhosherstov2021polyvit}. They deal with the problem of large input spaces by preprocessing the inputs into a small number of high-dimensional vectors using modality-specific operations or architectures. Perceivers~\cite{jaegle2021perceiver} can operate on raw inputs by adapting Transformers to operate on a latent bottleneck space, making it possible to deal with significantly larger input and output spaces and to build very deep models. Perceivers demonstrated competitive results in various domains, including audio, text and images~\cite{jaegle2021perceiverio}. However, they still require domain-specific preprocessing such as patching or convolutions for very large signals like high-resolution images or videos.

\noindent \textbf{Hierarchical models.} Unlike the equally popular ConvNets, most Transformer-based models are flat, having a constant number of activations and channels throughout.
Hierarchical variations have been recently proposed for NLP~\cite{dai2020funnel,nawrot2021hierarchical} and for vision, building upon ViT~\cite{fan2021multiscale,liu2021swin,ranftl2021vision}.
Here we propose a hierarchical version of Perceiver that reuses its technique of cross-attending to learned latents, but now also for adapting the internal resolution and number of channels between layers.

\noindent \textbf{Positional embeddings.} Transformers encode positional information by summing or concatenating specialized, hardcoded embeddings~\cite{vaswani2017attention, devlin2019bert, jaegle2021perceiver} (e.g. sinusoidal, Fourier, etc.), instead of defining and operating on local neighborhoods as in convolution-based architectures. While some prior works demonstrate learned embeddings on patched inputs (e.g.~\cite{dosovitskiy2020image}), to the best of our knowledge, scaling low-dimensional, completely-learned position embeddings to a million and more inputs has remained an open challenge.

\noindent \textbf{Masked auto-encoding.} The combination of masked auto-encoding (MAE)~\cite{vincent2008extracting} and Transformers has been tremendously successful in natural language processing, pioneered by famous models like BERT~\cite{devlin2019bert} and RoBERTa~\cite{liu2019roberta}. This success has recently started to percolate into specific perception domains: computer vision~\cite{he2021masked, bao2021beit}, point-cloud understanding~\cite{yan2022implicit}, and speech recognition~\cite{hsu2021hubert}. None of these previous works demonstrated, however, the impact of masked auto-encoding in learning high-resolution, modality-agnostic positional embeddings as done in this paper.

\noindent \textbf{Architectures for high-resolution data.} Modern architectures generally handle high-resolution data through convolutions and pooling (e.g.\ \cite{wang2020deep, ronnenberger2015convolutional, fischer2015flownet}), or through operations that are effectively a convolution (e.g.\ patching with linear projection, as is the case with~\cite{dosovitskiy2020image}). 
More recent architectures like Swin~\cite{liu2021swin} similarly implicitly rely on vision-specific image structure to perform windowing and merge operations.
HiP uses a non-convolutional approach to tackle high-resolution data that is more modality agnostic.

\section{Hierarchical Perceiver}
\label{architecture}

\begin{figure*}[t]
\centering
\includegraphics[width=0.55\textwidth]{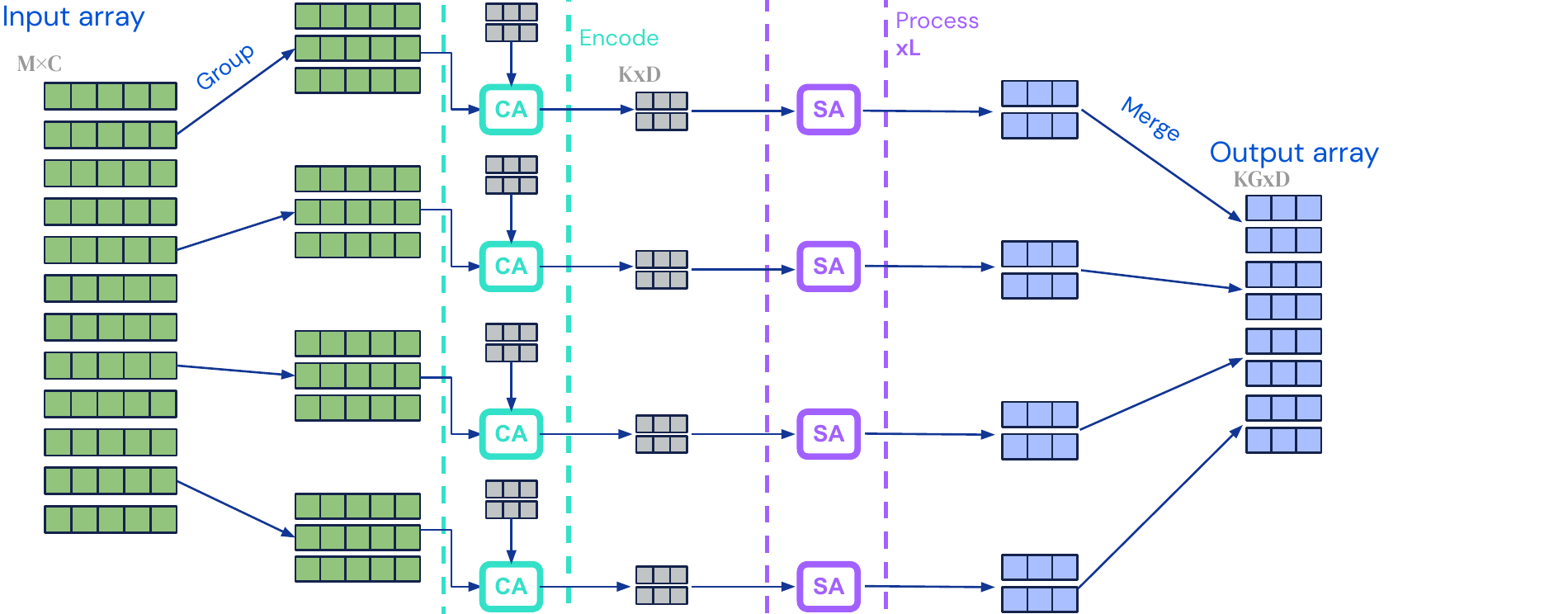}
\includegraphics[width=0.44\textwidth]{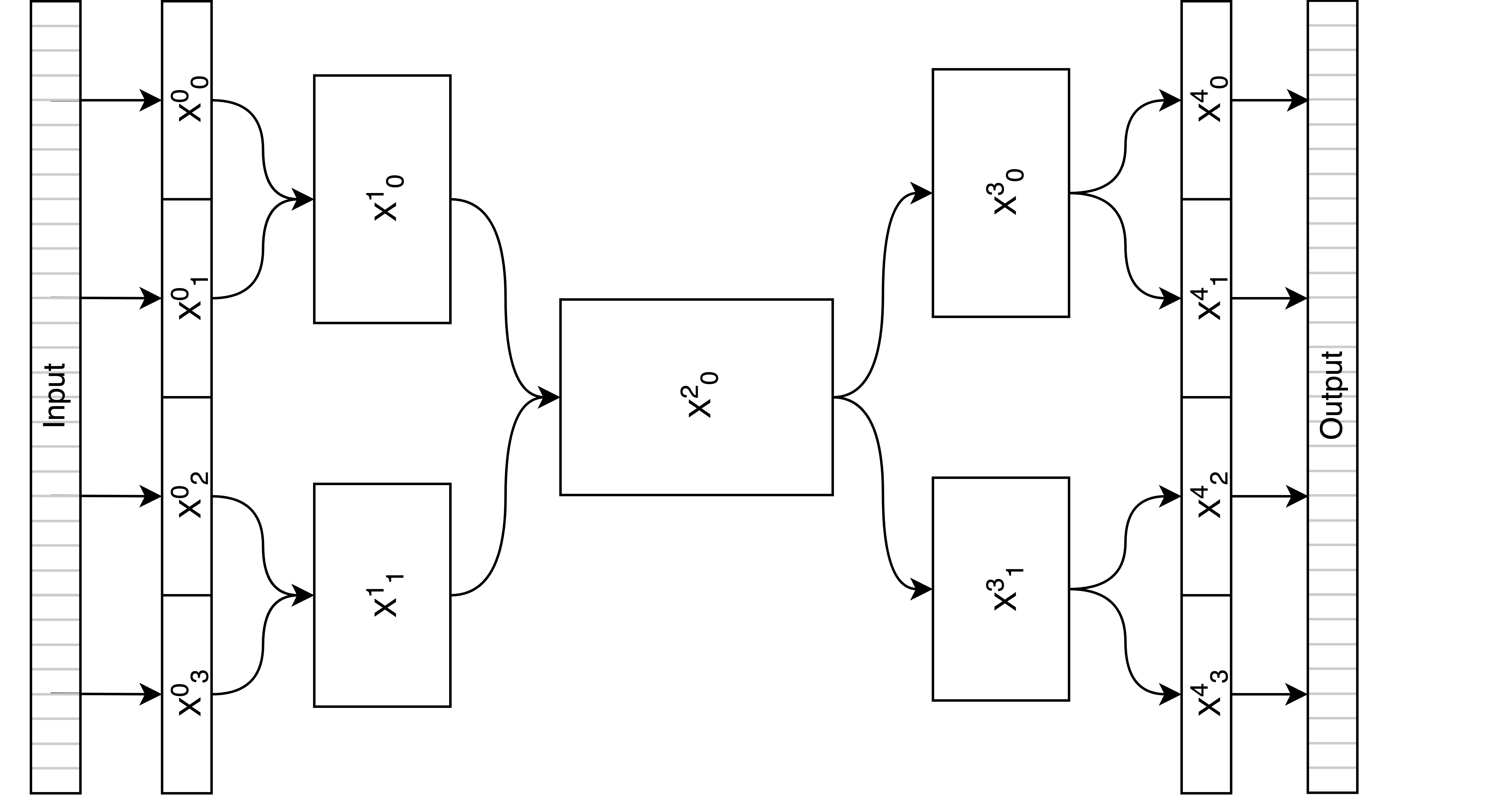}
\caption{\textbf{Left}: a depiction of the basic \ourmodel block structure. An input with $M$ tokens is grouped into $G$ groups, each $M/G$ in size. Within each group, a set of $K$ learned latent vectors $\Z_i$ (one for each group) cross attends to the group inputs $\X_i$. The output of the cross attention is then followed by a number of  self-attention + MLP blocks. The sets of output vectors from all groups $\Y_i$ are then merged together to form an output with $KG$ tokens. This can be used as input to the next block, forming a hierarchy. \textbf{Right:} An example of a 4-2-1-2-4 group Hierarchical Perceiver configuration - note that the internal block structure, as well as skip connections are omitted to make the hierarchy clearer. Many other configurations would be valid, e.g. 1-8-2-4. The width of different blocks illustrates that the number of channels is larger in layers closer to the model bottleneck.}
\label{fig:arch}
\end{figure*}

The Perceiver model assumes that input data $\X \in \mathbb{R}^{M\times C}$ always arrives as a set of $M$ tokens, each with $C$ channels. If there are different modalities and data sources, pre-processing steps such as flattening and padding may be required to arrive at this form.

The main trick Perceivers use to scale is introducing a set of learned latent vectors $\Z \in \mathbb{R}^{K\times D}$ which are updated by cross-attending over the inputs $\X$.
Expressed as a function of index dimensionality, the complexity of this input encoding process is $\mathcal{O}(MK)$ (compared with $\mathcal{O}(M^2)$ for Transformers).
The set of latents $\Z$ can be further processed using $L$ layers self-attention at complexity $\mathcal{O}(K^2)$ each, independently of input length.
The same scaling trick can also be used to efficiently produce outputs~\cite{jaegle2021perceiverio}.

\noindent \textbf{Introducing hierarchy.} Because Perceivers exclusively utilize Transformer-type operations, they are invariant to permutations of the input tokens and are hence agnostic to the structure of inputs before the flattening operation. Here we lean into the fact that flattening operations often preserve useful local structure for many modalities of interest such as images, video, audio and text. The \texttt{flatten} operation sequentially stacks elements from each index dimension (i.e. spatial or `token' dimensions), working backwards from the last index dimension until all elements have been stacked.\footnote{This corresponds to reshaping to 1D with NumPy's \texttt{reshape} operation~\cite{harris2020array}.} Because neighboring elements along an index dimension are often nearby in space or time, flattening preserves some (but not all) locality structure. See Figure~\ref{fig:flattening} for an illustration of how local structure is preserved when flattening an image. 

Note that neighboring elements after concatenation are not always nearby in space or time. For example, neighboring points after concatenation may belong to two separate rows of an image or may belong to the end of one modality array and the beginning of another modality array. This strategy is only able to partially preserve the locality structure present in input array's after flattening. For this reason, it represents one strategy for minimally preserving locality structure without re-designing the architecture for domains with new spatial, temporal, or modality structures. Motivated by these observations we allow our model to exploit locality by the introduction of \emph{groups}.

\begin{figure*}[t]
\centering
\includegraphics[width=14cm]{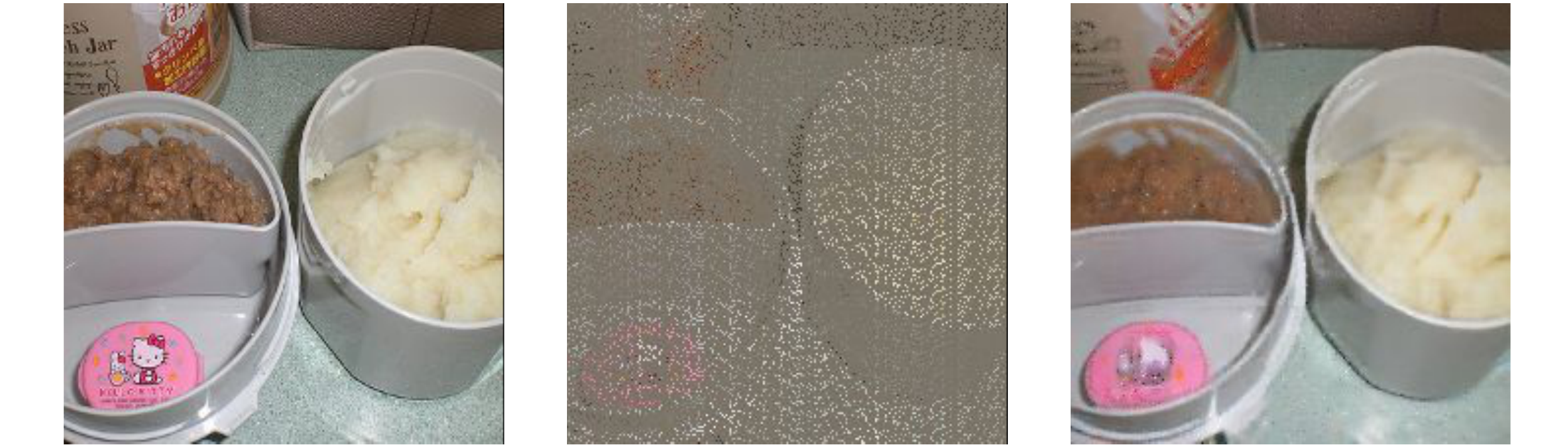}
\caption{Example masked auto-encoding results on one ImageNet image using 85\% masking rate for groupwise-masking. On the left we show the original image, in the middle the corresponding masked image and on the right the outputs of the 16-group model. Note that the masks were shared across groups (groups in HiP-16 for 224x224 images are sequences of 14 consecutive  pixel rows), and this is visible as a vertically recurring pattern.}
\label{fig:MAE}
\end{figure*}

\noindent \textbf{The basic \ourmodel block.} First, in order to utilize local structure to improve efficiency we \emph{group} the input to the block by splitting it into $G$ groups, each with $M/G$ tokens. Each of these groups is processed independently by a sequence of Perceiver-like operations: first cross attending within each group with a set of $K$ latent vectors (at $\mathcal{O}(MK/G)$ complexity) then applying several layers of Transformer-style self-attention and MLPs on the latents that result within each group (overall complexity $\mathcal{O}(G\,K^2$) per layer). This is done separately within each group separately but with all parameters shared, except for the learned latent vectors which are different and specific to each group.\footnote{This choice is motivated by the fact that in the multimodal case different groups may have different modalities so may benefit from distinct learned latent parameters.} The outputs from all groups are then \emph{merged} along the index dimension to produce the block output $\Y \in \mathbb{R}^{(KG)\times D}$ with $KG$ tokens, each with $D$ channels. This output can be used as input for the next block as explained below. See Figure \ref{fig:arch} for an illustration of the block structure.

We can organize blocks that follow this structure in a hierarchical manner where each block treats the previous block's output as input, grouping and processing accordingly. Figure \ref{fig:arch} shows an example of such a hierarchy. In order to produce dense outputs we again generalize Perceiver IO~\cite{jaegle2021perceiver} and construct a reverse hierarchy of layers for decoding. For classification we can cross-attend a learned query with the features directly from the bottleneck layer.

\noindent \textbf{Depth and width.} While Transformer-based models typically have a constant number of channels in all layers, several recent successful models, particularly in vision~\cite{fan2021multiscale,liu2021swin}, have adopted a ConvNet-type channel-depth progression -- using more channels deeper in the model where ``resolution'' is lower. This offers efficiency benefits and may provide a useful inductive bias. We follow this principle in \ourmodel as well.

\noindent \textbf{Skip connections.} We use sum-type skip connections between encoder and decoder before the cross-attention layers, inspired by encoder-decoder vision architectures such as Hourglass or U-Net~\cite{newell2016stacked,ronnenberger2015convolutional}, as these were shown to help with dense labeling tasks. We found them helpful for stabilizing training in masked auto-encoding -- especially in audio -- and for obtaining slightly lower losses. 

\subsection{Learned low-dimensional positional embeddings}

Models like ViT~\cite{dosovitskiy2020image} learn a separate positional embedding for each token (e.g.\ each patch) and combine it additively with the token representation. This is extremely general and flexible -- for images it was shown that it is not even necessary to feed the model xy coordinates; spatial information is inferred by training directly on the classification task. Learned embeddings work  well for ViT as the number of inputs is small, e.g.\ 256, but for Perceivers which can have orders of magnitude more inputs it works poorly~\cite{jaegle2021perceiver,jaegle2021perceiverio}. Instead, what works well with Perceivers is to concatenate hand-designed Fourier positional embeddings with every input.

\textbf{Downsides of Fourier embeddings.} There are two main issues with Fourier embeddings. First, they must be tailored to each modality, depending on whether the data is 1D, 2D, or 3D, and with the maximum frequency and number of frequency bands depending on properties of the input signal determined by human expertise or hyperparameter sweeps. The second problem is that they are large and become a memory bottleneck when trying to scale up models to high-resolution data. For example Perceiver IO used 64 frequency bands for ImageNet experiments, which corresponds to 258 fourier features. For a short 32-frames video at $224 \times 224$ resolution these positional embeddings alone would take 13GB of GPU or TPU memory, making it impractical to train models. ViT's learned embeddings are even higher-dimensional.

\begin{table*}[t!]
\centering
\resizebox{\textwidth}{!}{
\begin{tabular}{lccccccc|ccccccccccc}\toprule
Configuration           & \multicolumn{7}{c}{HiP-16} & \multicolumn{11}{c}{HiP-256}  \\
\hline
Input channels   & & & & 32 & & & & & & & & & 16 & & & & & \\
Groups   & 16 & 4 & 1 & 1 & 1 & 4 & 16 & 256 & 64 & 16 & 4 & 1 & 1 & 1 & 4 & 16 & 64 & 256 \\
Self-attention layers & 2 & 2 & 18 & 2 & 1 & 1 & 1 & 1 & 1 & 2 & 2 & 18 & 2 & 1 & 1 & 1 & 1 & 1\\
Heads    & 4 & 8 & 16 & 32 & 16 & 8 & 4 & 1 & 2 & 4 & 8 & 16 & 32 & 16 & 8 & 4 & 2 & 1 \\
Channels & 128 & 256 & 512 & 1024 & 512 & 256 & 128 & 64 & 96 & 128 & 256 & 512 & 1024 & 256 & 128 & 64 & 32 & 16   \\
Latent vectors per group & 128 & 256 & 256 & 64 & 256 & 256 & 128 & 32 & 64 & 128 & 256 & 256 & 64 & 256 & 256 & 128 & 64 & 32 \\
\bottomrule
\end{tabular}
}
\caption{Hyper-parameters defining the two architecture configurations considered in this paper: HiP-16 and HiP-256. Each number inside the brackets refers to a different block (e.g. block 0, block 1, etc.). See Figure~\ref{fig:arch} for a diagram showing the structure of a block as well as of an overall example configuration. In both cases the number of channels increases closer to the model bottleneck, the number of groups decreases down to 1 and most of the processing happens in the bottleneck -- we apply 18 self-attention layers there. The difference between both models is in the number of blocks; HiP-256 has two additional blocks on both encoder and decoder with 256 and 64 groups allowing it to process short raw videos. HiP-16 has only 16 groups in the first and last blocks.}
\label{tab:arch_config}
\vspace{-1ex}
\end{table*}

\textbf{Self-supervised learning to the rescue.} We will show in the experimental section that it is feasible to learn low-dimensional positional embeddings but that a plain classification loss is not enough to fully develop them. Instead, 
what is more effective is to train the models using random, uniform masked auto-encoding, before training on sparsely-supervised downstream tasks like classification. We use a separate linear layer for each input modality to map all inputs to the same dimensionality, then add the learned positional embeddings for every input, using either 16 or 32 dimensions.

In more detail: we randomly and uniformly mask a fixed percentage of the input tensor -- this is controlled by a hyperparameter. The input tensor can have multiple modalities and we mask them all uniformly with the same probability. We consider both the classic masking approach~\cite{devlin2019bert} of replacing the masked inputs with a learned token (\textit{uniform-masking}), and the more efficient alternative of dropping the masked inputs~\cite{he2021masked} -- we call this \textit{groupwise-masking}. Because of the model's group structure, to avoid poorly hardware-supported variable length inputs we subsample the same number of inputs in all groups (we sample a single random mask and replicate it for all groups). Groupwise-masking is complex to implement for multimodal inputs, so we use uniform masking in that setting. In both cases we query the final layer of the hierarchical decoder with the learned positional embeddings corresponding to masked inputs. Figure~\ref{fig:MAE} shows an example illustrating the masking pattern.

\textbf{Alternatives.} There are other ways to learn positional embeddings, for example it is popular to learn an MLP on xy coordinates for image applications -- this however still requires modality-by-modality design decisions and extracting such coordinates where relevant. In our experiments we found this possibility to still be inferior to Fourier features when trained just for classification. Another popular option is to use relative positional embeddings, but for Perceiver-type models this is not straightforward to do because inputs cross-attend to latent vectors instead of among themselves as more commonly done.

\subsection{HiP Configurations}

We propose two different HiP configurations and use them throughout the experiments. The first one, HiP-16 has slightly shallower hierarchical encoder / decoder modules than the second, HiP-256, otherwise both configurations are the same. They are fully described in Table~\ref{tab:arch_config}. The HiP-16 encoder has 16-4-1-1 groups for its 4 blocks (the decoder is symmetric) whereas the HiP-256 encoder has 256-64-16-4-1-1 groups. 

For problems with extremely many inputs, the input tensor size can dominate memory consumption, so for the more scalable HiP-256 we use just 16-dimensional positional embeddings (we also project the inputs to those dimensions because inputs and positional embeddings get summed); respectively 32-dimensional for HiP-16. In both cases we  privilege encoder over decoder depth and width because many of the tasks of interest require strong recognition capabilities.

\section{Experiments}

We start with ImageNet and compare the efficiency and accuracy of HiP and Perceiver IO. We also do most ablations and analysis on ImageNet.
Afterwards we cover the multimodal setting with raw audio+video on AudioSet, then experiment on a high-resolution vision task: semantic segmentation in PASCAL VOC. Finally we analyze the robustness of the model to point cloud data on ModelNet40. We use HiP-16 in all experiments except in AudioSet where the number of inputs / outputs is large -- we try only HiP-256 there. 

For masked auto-encoding pre-training we use the optimization hyperparameters from~\cite{he2021masked} and pre-train for 500 epochs with 85\% masking for most experiments because that seems to do slightly better than 75\% on ImageNet in preliminary experiments. On video datasets (AudioSet and Kinetics) we pre-trained for longer and found it helpful to use much higher masking rates. We show example masked auto-encoding validation curves in Appendix Figure~\ref{fig:mae_loss}. For finetuning we provide experimental details in each subsection. Other hyperparameters include using a 4x width expansion factor for MLPs in self-attention blocks, a 1x width expansion factor for cross-attention blocks and stochastic depth~\cite{huang2016deep} with 30\% drop rate in training.

\subsection{ImageNet}

\begin{table}
\centering
\begin{subtable}{}
\centering
\resizebox{0.45\linewidth}{!}{
\begin{tabular}{cccccc}\toprule
\textbf{Image res.} & 224 & 384 & 512 & 1024 & 2048 \\ \midrule
ResNet-50 & 46.0 & 20.7 & 12.5 & 3.4 & OOM \\ 
ResNet-101 & 30.8 & 14.2 & 8.7  & 2.4 & OOM \\  
VIT-B  & 14.9 &   5.7 &  2.8 & OOM  &   OOM \\
VIT-L  & 4.8  &  1.9  &   1.0 &    OOM  &   OOM \\
Perceiver IO        & 2.9 & 2.4 & 2.0 & OOM & OOM \\ 
\bottomrule
\end{tabular}}%
\end{subtable} \hspace{2mm}
\begin{subtable}{}
\centering
\resizebox{0.47\linewidth}{!}{
\begin{tabular}{cccccc}\toprule
\textbf{Image res.} & 224 & 384 & 512 & 1024 & 2048 \\ \midrule
HiP-16-Fourier      & 8.6 & 6.2 & 4.3 & OOM & OOM   \\ 
HiP-256-Fourier     & 6.7 & 4.6 & 4.2  & OOM & OOM \\  \midrule 
HiP-16              & 10.8 & 8.9 &  7.0  &  2.9  & OOM  \\ 
HiP-256             & 8.9 & 8.1 & 7.7 & 4.7 & 0.9\\ 
\bottomrule
\end{tabular}}%
\end{subtable}%
\vspace{2mm}
\caption{\label{tab:speed} Training steps per second (higher is better) for different input image resolutions (224x224, etc.) on TPUv3, using a batch size of 8 images per core. We compare HiP to the ImageNet version of Perceiver IO~\cite{jaegle2021perceiverio} for classification, without using the dense hierarchical decoder. OOM stands for ``out of memory'', so no timings could be recorded. Perceiver IO and HiP Fourier versions include the default 258-dimensional (64 bands) Fourier positional embeddings, whereas HiP-16 and HiP-256 use respectively 32d and 16d learned embeddings.}
\end{table}

We use standard 224x224 input crops for a total of 50,176 pixels. Unlike in masked auto-encoding, for classification experiments the decoder is unused. We finetune models similarly to prior work~\cite{liu2021swin} -- for 300 epochs, using 20 epochs of warmup, the AdamW optimizer~\cite{loshchilov2017decoupled} and a cosine decay schedule.

\noindent \textbf{Comparison to state-of-the-art.} Table~\ref{tab:imagenet_sota} compares our models with a (far from exhaustive) list of strong models. HiP-16 gets 81.0\% accuracy on ImageNet from pixels, up from 79.0 of Perceiver IO also from pixels. HiP-256 is only slightly worse at 79.9\% (local groups correspond to consecutive 196 row-wise pixels in this case which may not be ideal). When comparing to previous Perceiver models with learned position embeddings, accuracy is much higher. We report Perceiver IO results with MAE in the appendix. Finally, both HiP models are much faster than the Perceiver IO baseline, as shown in a side-by-side comparison in Table~\ref{tab:speed}. Overall accuracy is not far from top methods even though HiP does not use any 2D inductive biases (e.g. patching, convolutions).

\begin{table}
\centering
\begin{subtable}{}
\centering
\resizebox{0.45\linewidth}{!}{
\begin{tabular}{@{}lcccc@{}}
\toprule
\textbf{Model}           & Fourier & MAE & \textbf{Acc.} \\ \midrule
HiP-16 & \cmark &\cmark & 78.7 \\ 
HiP-16 & \cmark & \xmark & 78.8 \\
HiP-16 & \xmark & \xmark & 70.1 \\
HiP-256 & \cmark & \xmark & 76.9 \\
HiP-256 & \xmark & \xmark & 68.1  \\ \midrule
HiP-16 shuffled pixels & \xmark & \cmark & 76.3 \\
HiP-16 shuffled pixels & \xmark & \xmark & 68.8 \\ \midrule
HiP-16 & \xmark  & \cmark  & 81.0 \\  
HiP-256 & \xmark  & \cmark  & 79.9  \\ \midrule
\end{tabular}}
\hspace{4mm}
\end{subtable}%
\begin{subtable}{}
\centering
\resizebox{0.45\linewidth}{!}{
\begin{tabular}{@{}lcccc@{}}
\toprule
\textbf{Model}           & \textbf{Acc.} & \textbf{Params}         \\ \midrule
\textbf{ConvNet baselines} & & &  \\
ResNet-50~\cite{he2016deep}       & 78.6 & 26M         \\
NFNet-F6+SAM~\cite{brock2021high} & 86.5 & 438.4M         \\
\textbf{Transformer baselines} & & & & \\
ViT-B/16~\cite{dosovitskiy2020image} & 77.9 & 86M          \\
ViT-B/16~\cite{he2021masked} & 82.3 & 86M          \\
ViT-B/16 + MAE~\cite{he2021masked} & 83.6 & 86M          \\
Swin-B~\cite{liu2021swin}  & 83.5 & 88M  \\  \midrule
\textbf{w/ 2D Fourier features} & & & \\
Perceiver~\cite{jaegle2021perceiver} & 78.6 &  42.1M           \\ 
Perceiver IO~\cite{jaegle2021perceiverio} & 79.0 & 48.4M          \\
\textbf{w/ learned position features} & & & & \\
Perceiver (learned pos)~\cite{jaegle2021perceiver} & 67.6 & 55.9M          \\
Perceiver IO (learned pos)~\cite{jaegle2021perceiverio} & 72.7 & 62.3M   \\ 
HiP-16 & 81.0 & 97.9M\\  
HiP-256 & 79.9 & 102.3M \\ \midrule
\end{tabular}}
\end{subtable}
\caption{\label{tab:imagenet_analysis}\label{tab:imagenet_sota} \small \textbf{Left:} analysis of the impact of different factors on HiP ImageNet performance (top-1 accuracy, higher is better). \textbf{Right:} results on ImageNet image classification (top-1 accuracy, higher is better). Results for Perceiver / Perceiver IO baselines are taken from the respective papers~\cite{jaegle2021perceiver,jaegle2021perceiverio} and are from raw pixels -- without patching or convolutions.}
\end{table}

\noindent \textbf{Ablations: MAE vs no MAE.} Table~\ref{tab:imagenet_analysis} compiles several results that aim to shed light into what is important in HiP. We first study the importance of masked auto-encoding (MAE): when using learned positional embeddings, performance is dramatically lower without MAE. MAE pre-training for a model using Fourier positional embeddings however leads to similar performance which may suggest that the main value of MAE in our case is to obtain better low-dimensional positional embeddings. We experimented also masking random 75\% of 16x16 patches~\cite{he2021masked} instead of pixels (while still operating on pixels) and finetuning performance was about 2\% worse. Compared to previous work we did not try layerwise learning rate decay in finetuning~\cite{he2021masked} which may boost performance further.

\noindent \textbf{Ablations: importance of local structure.}
We also evaluate the importance of local structure being preserved in the way data is presented to the model by
destroying this structure with a fixed shuffling of the pixels.
Table~\ref{tab:imagenet_analysis} shows that performance drops by 5\% with MAE and by 12\% without MAE.
Spatial coordinates are never input to the model in training or testing; the accuracy drop is due to the impossibility of the model comparing spatially nearby pixels directly in early layers as the shuffling assigns them to separate groups.
With MAE the model may perhaps be more capable of preserving individual pixel information into deeper layers. 

\begin{figure*}[t]
\centering
\includegraphics[width=6.5cm]{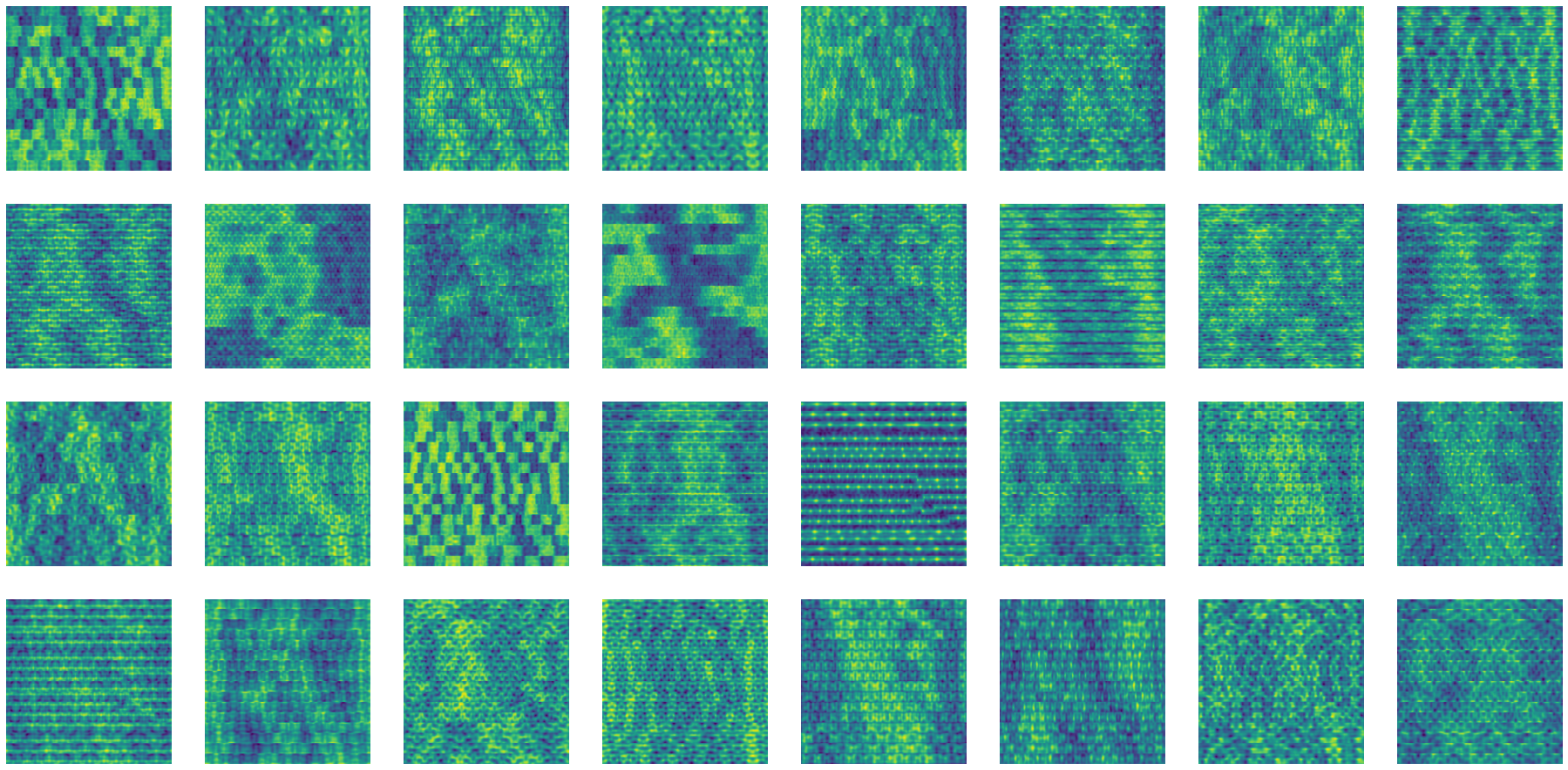}\hspace{5mm}
\includegraphics[width=6.5cm]{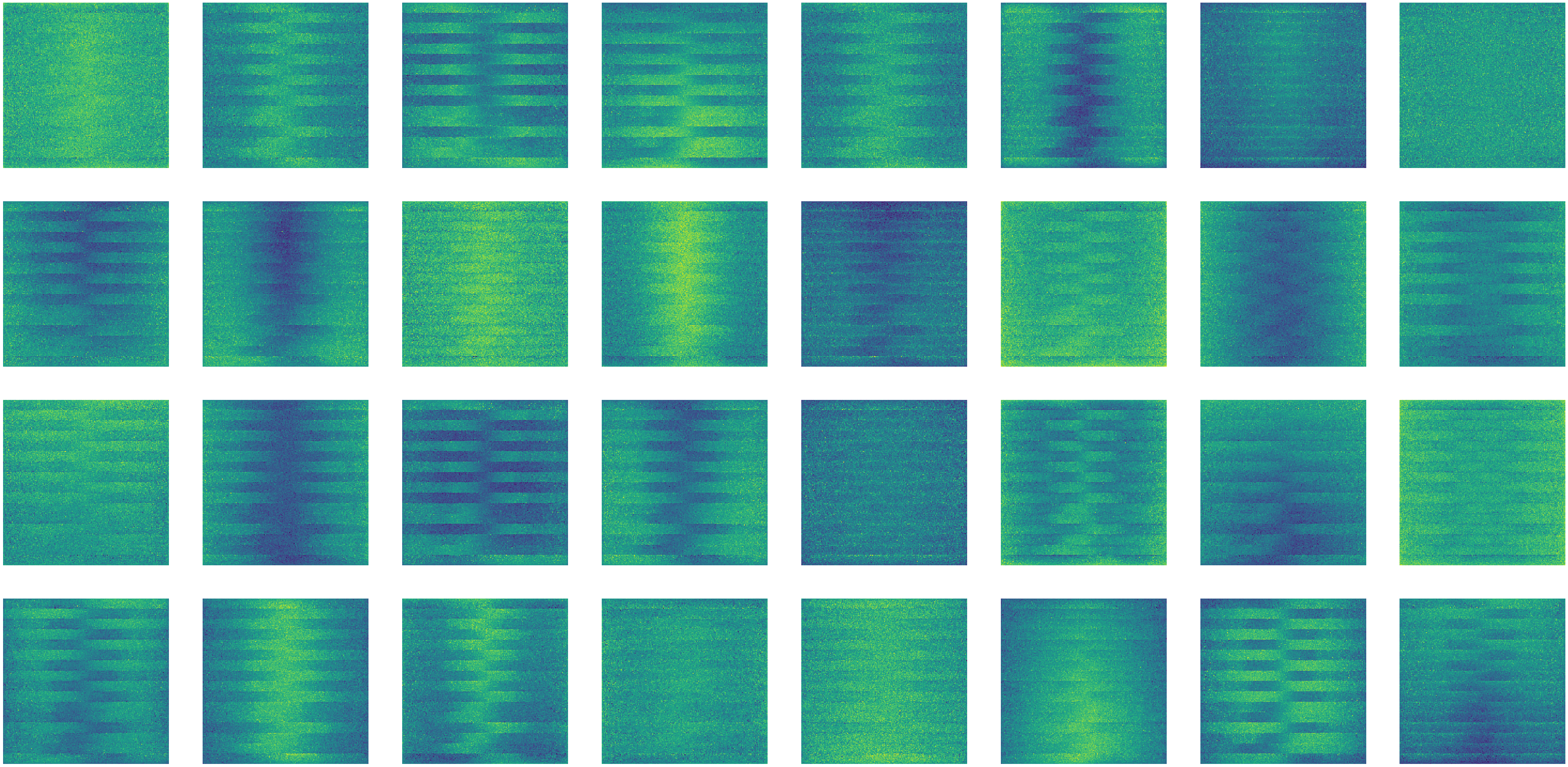}
\caption{HiP-16 list of 32 positional embedding channels, with (left) and without (right) MAE pre-training, in both cases followed by classification finetuning, both on ImageNet. MAE leads to rich low-dimensional positional embeddings  that are competitive with high-dimensional hand-designed Fourier embeddings for various downstream tasks. Each image displayed is a single channel of the learned embedding. Without this pre-training the resulting learned embeddings obtain very poor downstream performance. Each of these small images is 224x224 resolution, so this is better seen on a computer screen while zooming significantly in.}
\label{fig:pos_embs}
\end{figure*}

\noindent \textbf{Positional embedding analysis.}
We visualize positional embeddings after finetuning HiP-16 for classification, with / without masked auto-encoding pretraining in Figure~\ref{fig:pos_embs}.
HiP-16 uses 32d positional embedding channels and divides the flattened inputs into 16 groups that correspond to sequences of 14 consecutive pixel rows.
This grouping is visible in the resulting embeddings as 16 horizontal stripes.
We show additional visualizations in Appendix Figures~\ref{fig:distances_pos_emb} and~\ref{fig:pca_pos_emb}.

\subsection{AudioSet}
In this section we pre-train and fine-tune HiP on audio-visual inputs. This domain is especially challenging for its high dimensionality (both waveform and images are directly fed to the model without pre-processing). In particular, the input is a concatenation of the raw audio and the pixels. We use $8$ $224 \times 224$ resolution video frames (sampled at $3.125$fps) and the corresponding $2.56$s of audio (sampled at $48$kHz). This results in $401$k input samples for video and  $123$k samples for the audio. In the next subsection we report results on video classification in Kinetics~\cite{carreira2018short}, using 2.5s clips at high frame-rates (12.5 fps) for a total of $1.6M$ inputs and show models still train well. To process this very high-dimensional input we use the HiP-256 architecture and pre-train our models using the masked auto-encoding reconstruction loss on the flattened video+audio signal. These experiments aim to challenge HiP with a large reconstruction problem -- note that spatial and temporal coordinates are never provided for any of the inputs and must be inferred. Following~\cite{jaegle2021perceiver}, we use the AudioSet dataset~\cite{gemmeke2017audio} which has $1,879,569$ clips (audio+frames) for training and $18,427$ in the test set (at the time of submission), which are annotated with a multi-label hierarchy consisting of $527$ classes. We fine-tune our pre-trained HiP models on audio-visual multi-label classification of the $527$ AudioSet classes using a sigmoid cross-entropy loss. 

The top-performing methods on this dataset preprocess audio into visual representations: spectrograms and often pre-train on ImageNet~\cite{nagrani2021attention}. In Table~\ref{tab:audioset} we compare to the top methods operating on raw waveforms: VATT~\cite{akbari2021vatt} and Perceiver (we believe the latter is the only other method reporting results on raw audio + video). For fine-tuning and evaluation we use a similar protocol as in~\cite{jaegle2021perceiver}.  In both cases our model outperforms the baselines, which ingest pre-processed tokens (128-d ``patches'' of raw audio), but does so more markedly in the audio-only case. The larger improvement with audio-only may be because audio is the more informative modality in this dataset and the model devotes more its early capacity to video, since groups are fairly distributed -- video has more inputs so gets assigned more groups. We fine-tune for $300$ epochs using the AdamW optimiser~\cite{loshchilov2017decoupled}, with a batch size of $256$ and a cosine learning rate schedule with maximum value of $3 \times 10^{-5}$ and $5$ epochs of warmup. We experimented with longer masked auto-encoding pre-training and higher masking rates on AudioSet: for audio-only, initial models pre-trained for 300 epochs using 85\% masking did about 2.5\% absolute worse than the version with 1000 epochs and 90\% masking shown in the table. For audio+video we obtained best results using even higher masking: 97.5\%, also for 1000 epochs. It is possible that different masking rates for each modality may help further, but we have not tried it.

\subsection{Kinetics-600 experiments}

We experimented with HiP-256 on Kinetics-600 classification. The inputs in this case are 32-frame 224x224 resolution clips at 12.5fps. The full videos are up to 10s long and we resize them to a short side of 256 pixels. We follow the usual practice of evaluating on multiple clips on the validation set for evaluation, in this case 8x3 clips (8 equally spaced slices in time, 3 slices in space). 

We pre-trained the models using MAE with uniform masking on the Kinetics-600 training set for either 1000 or 2000 epochs and finetuned them for 100 epochs on that same data, then evaluated on the validation set. We used the same hyperparameters as for AudioSet otherwise, but did not use mixup augmentation in finetuning (so only random cropping, flipping, random resize, and color augmentation). Results are shown in table~\ref{tab:kinetics}. Note that transformer-based video models typically rely on ImageNet pretraining or otherwise from scratch they produce poor results in Kinetics, to the point where results are not even included in the papers (e.g. ViVIT~\cite{arnab2021vivit}, TimesFormer~\cite{bertasius2021space}). A notable exception is Multiscale VIT~\cite{fan2021multiscale}, which however uses more more extensive augmentation and generally a better finetuning setup. Using pixelwise MAE we get decent results with HiP-256 without pre-training on other datasets.

\begin{table}[b]
\centering
\begin{tabular}{@{}lcccc@{}}
\toprule
\textbf{\# epochs pre-training} & \textbf{Masking \%} & \textbf{\#inputs} & \textbf{Finetuning Top-1 accuracy}           \\ \midrule
1000 & 97.5\% & 1,605,632 & 69.2\% \\
1000 & 99.375\% & 1,605,632 & 74.5\% \\ 
1000 & 99.685\% & 1,605,632 & 69.1\%  \\ \midrule
2000 & 98.75\% & 1,605,632 & 72.8\%  \\ 
2000 & 99.375\% & 1,605,632 & \textbf{75\%}  \\ 
2000 & 99.685\% & 1,605,632 & 71.7\%  \\ \bottomrule
\end{tabular}
\hspace{0.5mm}
\vspace{2mm}
\caption{\label{tab:kinetics} Classification results on Kinetics 600~\cite{carreira2018short}. We found that extremely high level of masking is required for best performance with high frame-rate video, perhaps because of the high-redundancy of the signal. The optimal amount of masking seems to be around 99.375\% (one in every 160 pixels is visible, roughly one pixel in each 6x5x5 space-time volume).}
\end{table}

\subsection{PASCAL VOC}
We also explore HiP for semantic segmentation, a dense visual task that aims to identify the object classes for each image pixel. Following prior literature \cite{byol, moco}, we pre-train both HiP and ResNet-50 on ImageNet. The pre-training procedure (masked auto-encoding followed by supervised fine-tuning) described in Section~\ref{architecture} is used to pre-train HiP. We use the PASCAL VOC 2012 dataset, fine-tuning on the commonly used
\texttt{train\_aug2012} partition and report the mean intersection over union (mIoU) on the \texttt{val2012} set.

To perform segmentation with HiP, we query the decoder using the fully learnt position embeddings to determine the class of each pixel. We either use the proposed hierarchical decoder (`full'), or add a single-layer cross-attention decoder onto the bottleneck latents in HiP (cutting the HiP architecture in half, discarding the original decoder layers), similar to what we do for classification. Our ResNet-50 baseline uses the same fully convolutional segmentation head as used in \cite{byol, moco}. While the original Perceiver IO did not demonstrate results on segmentation~\cite{jaegle2021perceiverio}, it can be easily applied for this task. The best ImageNet from-pixels Perceiver IO model~\cite{jaegle2021perceiverio} uses 8 repeated blocks, each with 6 self-attends (parameters shared between blocks).
We also pre-train a leaner version of Perceiver IO (shallow) with better speed on ImageNet; this adapted model has just 1 block with one cross-attend and 12 self-attends. These Perceiver IO baselines use fixed Fourier position embeddings with 64 bands as in the original paper.

Table~\ref{tab:segmentation_results} shows the mIoU and training speed (in training steps/second, with fixed batch size) of three reference models and two HiP-16 variants. Sample segmentation outputs from HiP-16 are shown in Appendix Figure~\ref{fig:pascal_segmentations}.
Following common protocol, we use standard 512x512 resolution input, and train each model across 8 TPU-v3 cores.
Both versions of HiP obtain mIoU on par with our optimized, convolutional ResNet-50 baseline, but are slightly slower than the ResNet-50.
With its fewer layers, HiP (bottleneck) runs slightly faster than HiP (full), but this speed comes at the cost of 1 mIoU.
Both HiP variants outperform our PerceiverIO baselines in both accuracy and training speed.
We note that the original PerceiverIO (denoted with a *), may be underperforming (mIoU) because this very deep model overfits on the small PASCAL VOC dataset.

\subsection{ModelNet40}

As a last experiment we consider shape classification on ModelNet40~\cite{wu20153d}, a dataset of point clouds with 2048 elements, each with 3 spatial coordinates (x, y, and z) and 40 shape classes. The challenge here for architectures with locality assumptions is that points are randomly permuted.
We do not augment the data, nor do we employ specialized geometrical representations (these are quite important for driving performance in state-of-the-art methods, such as~\cite{xiang2021walk}), so we compare only to our own baseline: a Perceiver IO model operating in the exact same experimental setting.
We sweep over the number of self-attention layers (8, 12, 16), channels (512, 1024) and latents (256, 512, 1024) for this baseline, and for both models we sweep learning rate and weight decay -- we report the best numbers across all sweeps.
HiP-16 is used without modification.

Table~\ref{tab:modelnet40} shows that HiP gets very slightly worse performance than Perceiver IO when training from scratch, possibly because the permutation invariance of Perceiver IO gives it a small edge. However, after MAE pre-training HiP-16 slightly outperforms Perceiver IO.
The reason for this is not completely clear. 
We inspect the positional embeddings before and after MAE pre-training and they looked similar: no visible structure as expected, as the points are randomly permuted. Pre-training may provide regularization  on this small dataset (9,843 training examples).

\begin{table}
\centering
\begin{subtable}{}
\resizebox{0.35\linewidth}{!}{
\begin{tabular}{@{}lccc@{}}
\toprule
\textbf{Model} & \textbf{Modalities} & \textbf{\#inputs} & \textbf{mAP}           \\ \midrule
Perceiver~\cite{jaegle2021perceiver} & A+V & 13,024 & 43.5  \\
HiP-256 & A+V & 524,288 & \textbf{43.8}  \\ \midrule
VATT~\cite{akbari2021vatt} & A & 375 & 39.4 \\ 
Perceiver~\cite{jaegle2021perceiver} & A & 480 & 38.3 \\
HiP-256 & A & 122,880 & \textbf{41.3}  \\ \bottomrule
\end{tabular}}
\end{subtable}%
\hspace{0.5mm}
\begin{subtable}{}
\centering
\resizebox{0.30\linewidth}{!}{
\begin{tabular}{@{}lcc@{}}
\toprule
\textbf{Model}           & \textbf{mIoU} & \textbf{Steps/Sec}         \\ \midrule
ResNet-50~\cite{he2016deep}               & 70.5 & 14.8     \\
Perceiver IO~\cite{jaegle2021perceiverio}  & 62.0  & 7.2      \\
Perceiver IO (shallow)                      & 67.7 & 9.9      \\
HiP-16 (bottleneck)                           & 70.0 & 12.7     \\
HiP-16 (full)                                 & \textbf{71.0} & 11.4     \\
\bottomrule
\end{tabular}}
\end{subtable}%
\hspace{0.5mm}
\begin{subtable}{}
\centering
\resizebox{0.15\linewidth}{!}{
\begin{tabular}{@{}lc@{}}
\toprule
\textbf{Model}           & \textbf{Acc.}       \\ \midrule
Perceiver IO & 77.4  \\
HiP-16 (no MAE) & 75.9 \\ 
HiP-16 (with MAE) & \textbf{80.6} \\\midrule
\end{tabular}}
\end{subtable}
\vspace{2mm}
\caption{\label{tab:segmentation_results} \label{tab:audioset} \label{tab:modelnet40}  \small  \textbf{Left:} Classification mean average precision (mAP) on AudioSet using raw audio (A) and video (V).  The Perceiver and VATT baselines  preprocessed the inputs into audio and video patches (e.g. by concatenating pixels within each 2x8x8 space-time volume into a single vector). With HiP we directly feed in individual pixels and raw audio magnitudes. \textbf{Middle:} Results on PASCAL VOC semantic segmentation. \textbf{Right:} performance on ModelNet40 point cloud classification (top-1 accuracy, higher is better).}
\end{table}

\section{Discussion}
\label{sec:discussion}

\noindent \textbf{Connectivity learning.} 
Perceiver and Transformer models can in principle learn any connectivity pattern in a soft way as they perform global attention.
However, dense connectivity hinders scaling them to many inputs.
Convolutional networks employ extremely (hard) local operations and are more scalable, but their local neighborhoods are hand-designed and must be tailored for different modalities or combinations thereof.
The proposed HiP model bridges both families of models, enforcing a very loose, modality-agnostic local connectivity (that hinges on the assumption that the data exhibits explicit locality in the way it is presented, e.g.\ pixels not being randomly permuted).
HiP still offers significant room for soft-connectivity learning  as it uses few local neighborhoods (groups).
Most of its processing is still fully global and happens at HiP's bottleneck.

\noindent \textbf{The need for self-supervised pre-training.} One of the main results of this paper is that positional embeddings cannot be learned for high-resolution signals using global classification losses alone, which is not the case for low-resolution signals. Are unshared positional embeddings inadequate for high-resolution signals? or is classification simply poorly suited for learning high-resolution position features in general? We showed that dense masked auto-encoding is sufficient to learn good positional embeddings: could other approaches like regular auto-encoding, or (dense) contrastive learning approaches lead to results of similar quality? We leave these questions for future work.

\noindent \textbf{Many small inputs.} We want models that can take in arbitrary modalities. Some modalities will be low-dimensional but high-resolution, others high-dimensional but low-resolution. HiP  projects them all to latents of the same dimensionality. We think it may be better to operate on low-dimensional inputs, and break up high-dimensional ones into multiple smaller vectors, glued together by appropriate per-example learned embeddings. A more flexible solution may lead to better results, but we leave this for future work.

\noindent \textbf{Augmentation: the last bastion of domain knowledge?} We demonstrated competitive results with the same model on various modalities with minimal preprocessing and prior knowledge -- nowhere did we employ patching, convolutions, spectrograms, etc., even for very large inputs -- only raw data.
This is the setting that may be most promising for AutoML or \textit{anymodal} learning.
We still exploited domain knowledge in the set of augmentations employed for each experiment, which may be necessary because the datasets were insufficiently large for anymodal models. We suspect it will be possible to push anymodal models even further without augmentation by training on larger datasets and developing better pretraining procedures.

\bibliography{bib}
\bibliographystyle{unsrt}

\newpage
\appendix
\onecolumn
\section{HiP: Hierarchical Perceiver -- Appendix}

We include in this appendix various additional experiments and visualizations promised in the main body of the paper.

\subsection{Perceiver IO + MAE}

We experimented with MAE pre-training of Perceiver IO. Again we used uniform masking and the same pre-training and finetuning hyperparameters as for the other ImageNet experiments.

We set up a Perceiver IO model with 16 self-attention layers, 1024 channels and 512 latent vectors. We project pixels to 256-D using a learned linear layer and sum them with learned positional embeddings of the same size. A difference to the Perceiver IO paper is the use of stochastic depth regularization and longer finetuning(300 epochs, in line with the rest of our experiments). Similar to HiP we observe large improvement from random pixel MAE from 76\% to 81.5\%. The Fourier version gets 80.3\%. Note however that the model is 2x slower than HiP-16 on ImageNet and less scalable to large inputs than our models.

\subsection{Additional positional embedding visualization}

We include more analysis of the learned HiP-16 positional embeddings in this section. 

\begin{figure*}[t]
\centering
\includegraphics[width=\textwidth]{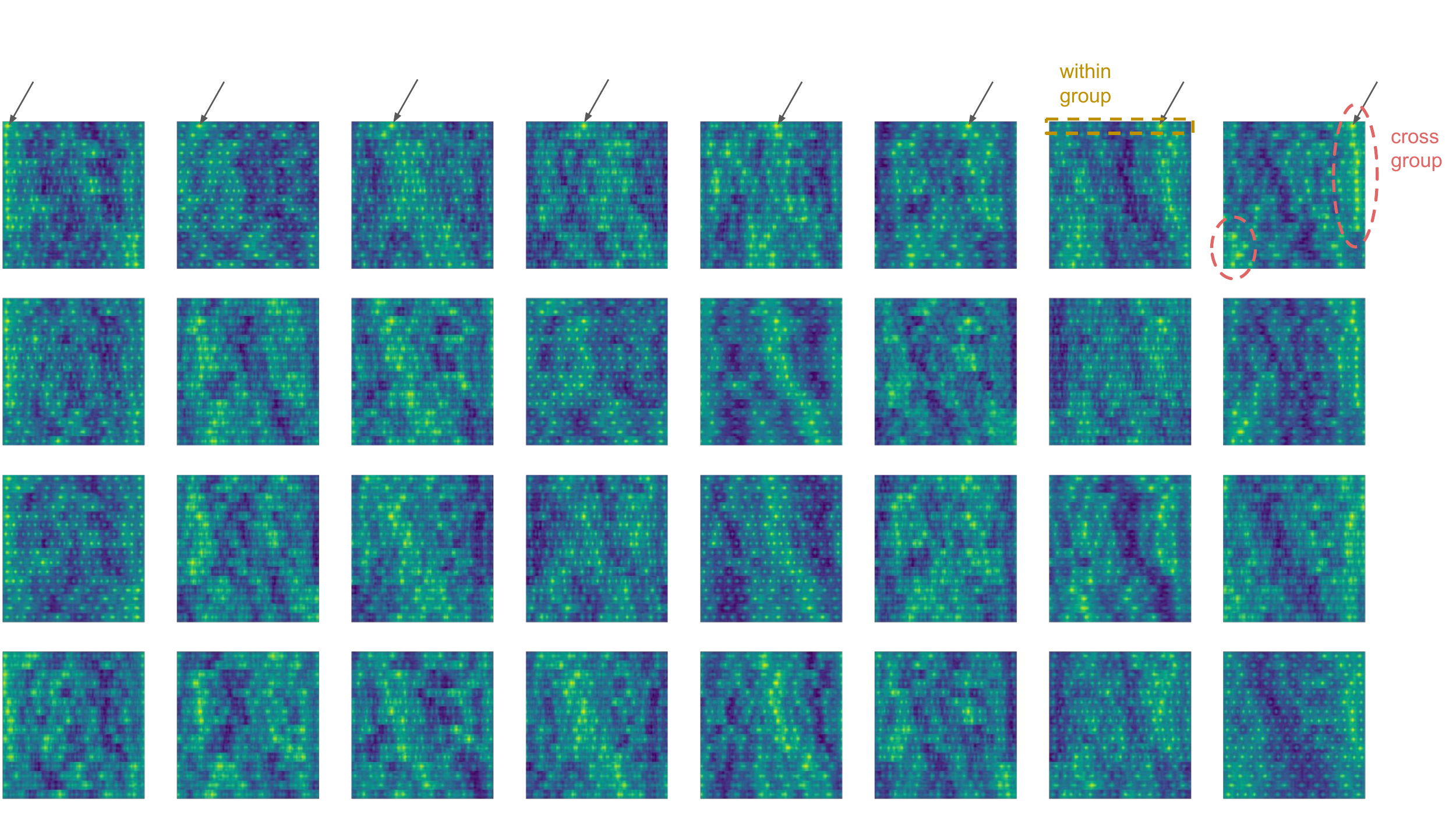}\hspace{11mm}
\caption{Visualizing the inner product distance between learned positional embeddings within and across groups for a HiP-16 model trained on ImageNet. Each image corresponds to a different positional embedding channel (there are 32 channels in total, hence 32 images are shown here). Images show the inner product distance between a single pixel (taken 14 pixels apart, across the image, marked in black arrows for the first row) and all other pixels. As can be seen, learned embeddings are close for nearby pixels within each group, as well as for similar locations across different groups (up to phase). Note that each group in HiP-16 corresponds to a set  of 14 consecutive pixel rows, because images have 224 rows that divide evenly into 16 groups.}
\label{fig:distances_pos_emb}
\end{figure*}

Fig.~\ref{fig:distances_pos_emb} shows distances between the positional embeddings of different pixels. Generally, the learned embeddings seem to mostly serve the purpose of relating pixels within each group; this may be an effective strategy for obtaining compressed position codes as it obviates the need to avoid global collisions -- codes can repeat in other groups. It is possible that for establishing more global locations within the image the model uses instead the learned latent vectors associated with each group.

\begin{figure*}[t]
\centering
\includegraphics[width=10cm]{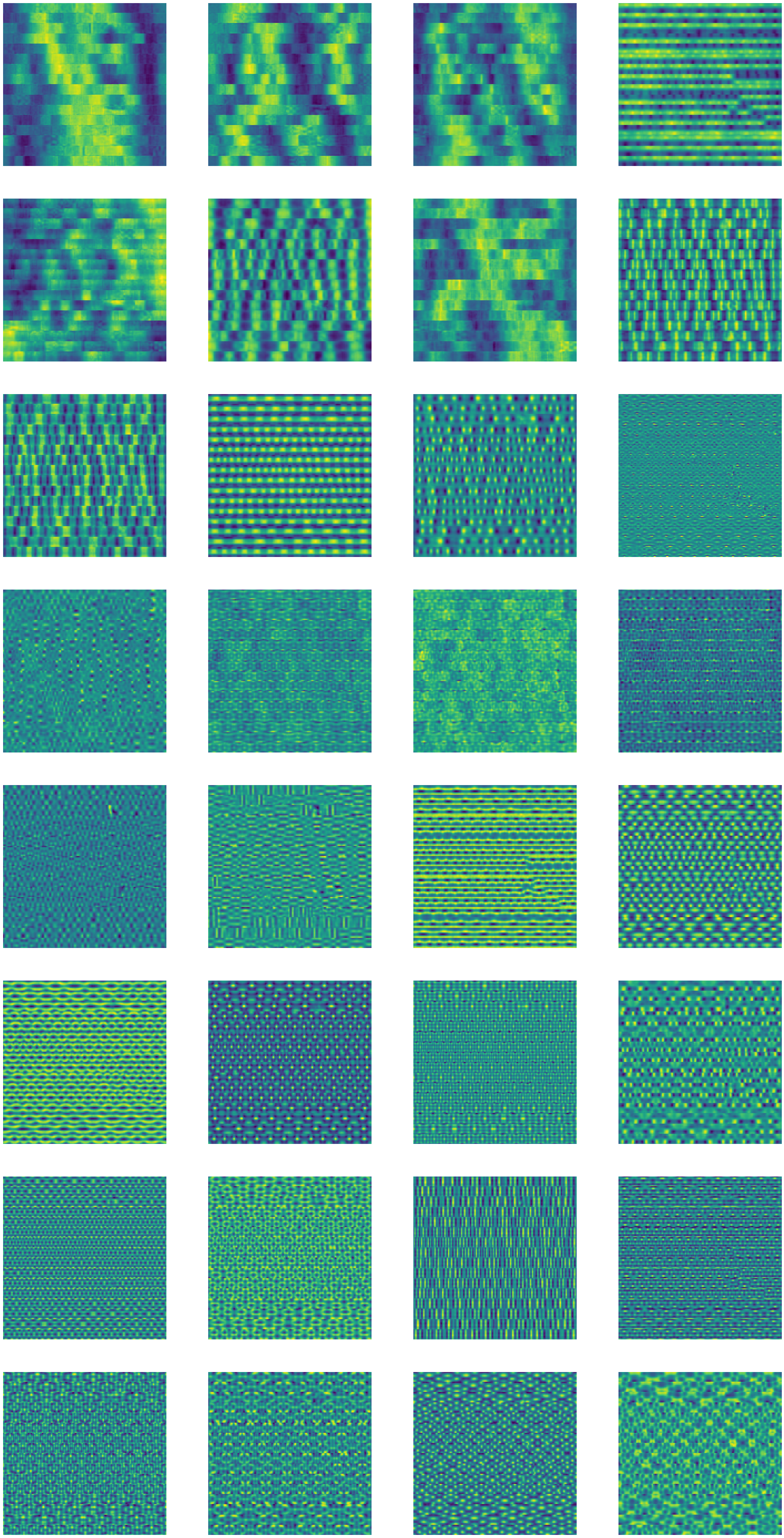}
\caption{Channel images of top PCA-projected spatial embeddings for HiP-16 trained on ImageNet, sorted left-to-right, top-to-bottom. As can be seen within each group (sets of 14 consecutive pixel rows), learned features resemble specific Fourier basis features.}
\label{fig:pca_pos_emb}
\end{figure*}

Fig.~\ref{fig:pca_pos_emb} shows a factorization of the 32d positional embeddings into their top principal components, which exhibit connections to Fourier basis.

\subsection{Masked auto-encoding validation loss curves}

\begin{figure*}[t]
\centering
\includegraphics[width=15cm]{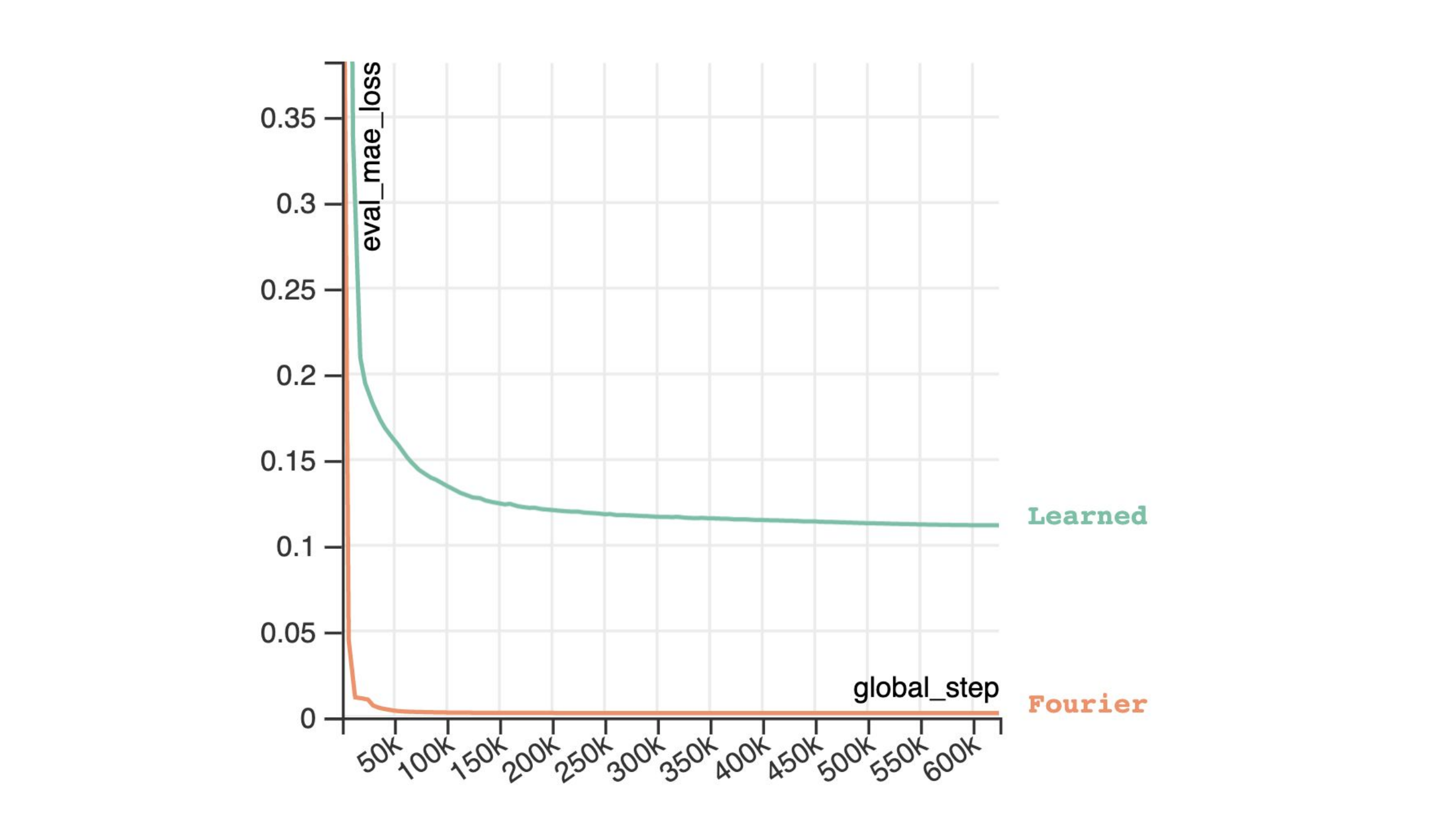}
\caption{Validation HiP-16 masked-autoencoding losses (mean-squared pixel error) using 85\% masking on ImageNet 224x224 images with either 32d "Learned" or 256d "Fourier" positional embeddings.}
\label{fig:mae_loss}
\end{figure*}

We show example curves for validation losses on ImageNet in fig.~\ref{fig:mae_loss}, using 85\% masking. Note that a version of HiP using Fourier positional embeddings gets considerably lower loss which suggests that they enable more precise pixel lookup than the low-dimensional learned ones. Results in the paper suggest that this level of precision may not be necessary for recognition and labeling tasks.

\subsection{Relationship to SWIN and ViT-based models}

ViT~\cite{dosovitskiy2020image} and more generally standard Transformers~\cite{vaswani2017attention} scale the amount of feature extraction they do by the input size -- e.g. if a ViT model processes 1024 patches, it will compute many more features than if it processes 256 patches. Perceivers do dominantly a fixed amount of computation, independent of the number of inputs as they operate in a fixed-size latent space. This makes it possible to do global computation efficiently in the  single-group processing bottleneck of HiP, independently of input resolution, unlike SWIN~\cite{liu2021swin}, where computation scales with input resolution. To deal with this SWIN  keeps computation local throughout the model, like ConvNets, and requires different "groups" within the network to communicate regularly, which is implemented via shifting operations. This is elegant but is not especially friendly for some types of hardware, like TPUs. We did take  inspiration from SWIN for some aspects, for example in choosing the pattern of number of self-attention layers and self-attention heads.

\subsection{Visualizing Sample Segmentations on PASCAL VOC}

\begin{figure*}[h]
\centering
\includegraphics[width=\textwidth]{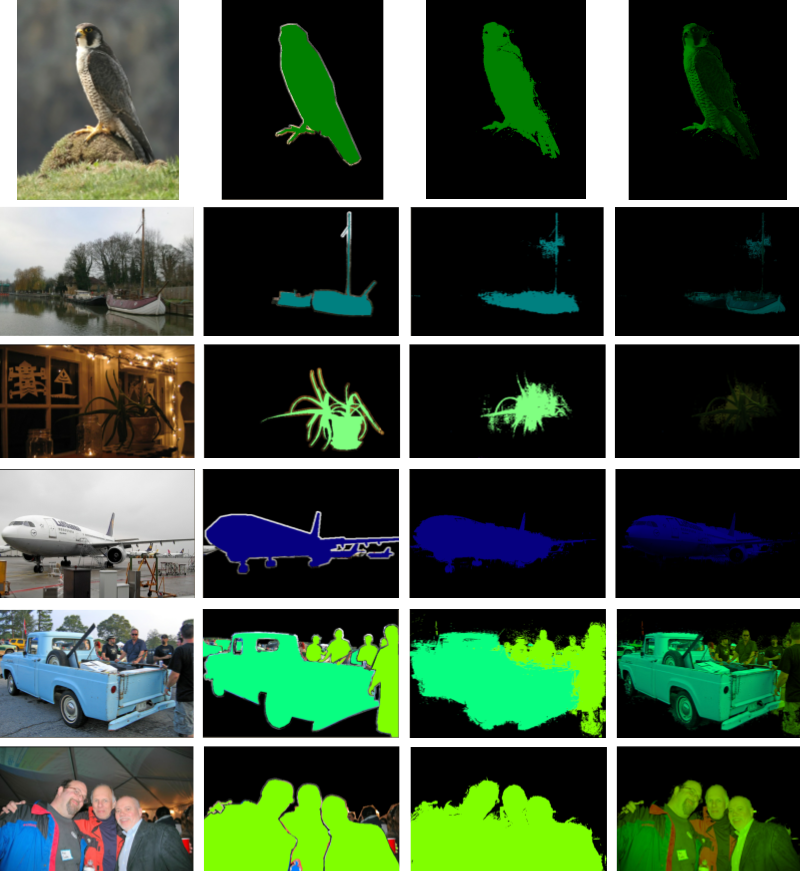}
\caption{Segmentation outputs of HiP-16 on randomly sampled PASCAL VOC 2012 images. \textit{Left column}: the original image. \textit{Middle left column}: the ground truth segmentation. \textit{Middle right column}: class-colored, HiP-16 segmentation prediction of the image. \textit{Right column}: the  HiP-16 segmentation prediction overlayed on top of the original image.}
\label{fig:pascal_segmentations}
\end{figure*}

Provided in Figure~\ref{fig:pascal_segmentations} are visualizations for the segmentation outputs of a HiP-16 on six randomly sampled images on the PASCAL VOC \texttt{val2012} set.

\end{document}